\pdfoutput=1
\documentclass[11pt]{article}

\usepackage{acl}

\usepackage{soul}
\usepackage{times}
\usepackage{multirow}
\usepackage{adjustbox}
\usepackage{latexsym}
\usepackage{changepage}
\usepackage{booktabs}
\usepackage{graphics}
\usepackage{graphicx}
\usepackage{blindtext}
\usepackage{float}
\usepackage{xcolor}

\usepackage[utf8]{inputenc}
\usepackage{amsmath}
\usepackage{changepage}
\newcommand{\paratopic}[1]{}
\newcommand{\stone}{\textsc{SToNe}\xspace}
\newcommand{\sen}{\textsc{SEN}\xspace}

\usepackage[T1]{fontenc}
\usepackage[utf8]{inputenc}

\usepackage{microtype}

\usepackage{inconsolata}

\title{LLMs for Targeted Sentiment in News Headlines: \\Exploring the Descriptive--Prescriptive Dilemma}

\author{Jana Juroš \and Laura Majer \and Jan Šnajder  \\ TakeLab, Faculty of Electrical Engineering and Computing, University of Zagreb \\ \texttt{\{jana.juros, laura.majer, jan.snajder\}@fer.hr} \\}

\begin{document}
\maketitle
\begin{abstract}
News headlines often evoke sentiment by intentionally portraying entities in particular ways, making targeted sentiment analysis (TSA) of headlines a worthwhile but difficult task. Due to its subjectivity, creating TSA datasets can involve various annotation paradigms, from \emph{descriptive} to \emph{prescriptive}, either encouraging or limiting subjectivity. LLMs are a good fit for TSA due to their broad linguistic and world knowledge and in-context learning abilities, yet their performance depends on prompt design. In this paper, we compare the accuracy of state-of-the-art LLMs and fine-tuned encoder models for TSA of news headlines using descriptive and prescriptive datasets across several languages. Exploring the descriptive--prescriptive continuum, we analyze how performance is affected by prompt prescriptiveness, ranging from plain zero-shot to elaborate few-shot prompts. Finally, we evaluate the ability of LLMs to quantify uncertainty via calibration error and comparison to human label variation. We find that LLMs outperform fine-tuned encoders on descriptive datasets, while calibration and F1-score generally improve with increased prescriptiveness, yet the optimal level varies.

\end{abstract}

\section{Introduction}

%News framing impacts information perception, shapes public opinion, and guides discussions on key topics \cite{framing_eu_politics}. 
News headlines -- succinct and attention-grabbing introductions to full news stories -- often evoke sentiment by portraying entities in specific ways. Targeted sentiment analysis (TSA) is the task of determining the polarity of sentiment expressed towards the target entity \cite{pei2019targeted}.  
While sentiment analysis is inherently challenging due to subjectivity, TSA introduces additional complexity by requiring the differentiation between targeted and overall sentiment. %\maybe{Analyzing the sentiment of news headlines is even more intricate for they are connotation-rich and culture-dependent, often involving implicit and indirect sentiment.} 
%\maybe{This requires a detailed interpretive effort, including interpreting quotes, identifying negative events linked to entities, and navigating mixed-view headlines \cite{hamborg_towards_2021}.}

For subjective tasks like TSA, the choice of data annotation paradigm is crucial. \citet{rottger-etal-2022-two} identified two contrasting paradigms: \emph{descriptive} and \emph{prescriptive}. The descriptive paradigm encourages subjectivity and diverse interpretations, typically with brief guidelines. In contrast, the prescriptive paradigm discourages subjectivity by providing detailed interpretation guidelines.

Fine-tuned encoders such as BERT \citep{devlin2019bert} show strong TSA performance across various languages \cite{wu_context-guided_2021,zhang_bert_2020, mutlu_dataset_2022}.  
However, using these models in different languages or domains requires new fine-tuning, and adapting them to low-resource languages necessitates pre-trained models and labeled data. In contrast, large language models (LLMs) offer a versatile approach to TSA across various domains by leveraging their broad linguistic and world knowledge, as well as in-context learning \citep{brown2020language}, without the need for annotated datasets or fine-tuning.
However, LLMs performance is often inconsistent and contingent on prompt design \citep{mizrahi2024state}, making it challenging to identify optimal settings. Furthermore, it is unclear how specific TSA criteria, defined during annotation, can be transferred using zero- and few-shot prompting.

In this paper, we compare the zero- and few-shot performance of open and closed-source LLMs to fine-tuned encoder models on datasets annotated following the descriptive or prescriptive paradigm. We then explore the influence of prompt design on the performance of LLMs for the prescriptive TSA dataset of Croatian news headlines. Similar to crafting effective annotation guidelines, finding the appropriate \emph{level of prescriptiveness} is essential in prompt design. The recent use of LLMs as data annotators \citep{wang-etal-2021-want-reduce,DBLP:journals/corr/abs-2306-00176,alizadeh2023opensource} further invites a direct comparison of annotation paradigms and prompt design: less prescriptive prompts give more interpretive freedom, while highly detailed prompts constrain it.
Building on this parallel, we evaluate the predictive accuracy of LLMs using prompts constructed from annotation guidelines with different levels of prescriptiveness, ranging from plain zero-shot to elaborate few-shot prompts matching annotation guidelines. 

%, drawing parallels with annotation paradigms for subjective tasks. 

%Drawing from the descriptive-prescriptive dichotomy in linguistics and ethics, \citet{rottger-etal-2022-two} proposed two contrasting paradigms for data annotation: the \emph{descriptive} paradigm, which fosters annotator subjectivity and encourages diverse interpretations, and the \emph{prescriptive} paradigm, which discourages subjectivity and mandates adherence to specific interpretation guidelines. 

%The recent utilization of LLMs as data annotators \citep{wang-etal-2021-want-reduce,DBLP:journals/corr/abs-2306-00176,alizadeh2023opensource} invites a direct comparison of these paradigms and prompt design: a simple, less prescriptive prompt grants the LLM more freedom in interpreting the input, whereas a detailed, highly prescriptive prompt restricts the interpretation.

Another interesting connection between annotation and prompting is label variation. Regardless of whether subjectivity is encouraged, some human label variation is inevitable in subjective tasks and may be leveraged to improve model performance \cite{mostafazadeh_davani_dealing_2022}. Similarly, LLM inconsistency, typically viewed as a limitation, can diversify responses to emulate human label variation. Recent LLM uncertainty quantification methods \cite{rivera_combining_2024, xiong_can_2023, tian_just_2023} can be used for the same purpose. Building on this idea, we assess LLMs' capability to quantify predictive uncertainty in TSA of headlines using calibration error and compare label distribution with human label variation.

Our experiments mainly focus on a Croatian dataset labeled with TSA on news headlines accompanied by detailed, prescriptive annotation guidelines. Additionally, we evaluate zero-shot LLMs and BERT on English, Polish, and Spanish TSA datasets with less prescriptive guidelines. 
Our contributions include (1) comparing LLMs and BERT for TSA on news headlines in four languages, (2) evaluating the effect of prompt prescriptiveness on LLMs' predictive accuracy, and (3) assessing calibration error and label distribution across models based on prompt prescriptiveness. This study offers valuable insights into LLMs' zero- and few-shot potential for TSA of news headlines.

\section{Related Work}

\paratopic {SA općenito u news i news headlines.}

Sentiment analysis of news headlines is an important task that has garnered significant attention in prior work \citep{agarwal_opinion_2016,stock2016,aslam2020sentiments,stock2021,rozado_longitudinal_2022}. In addition to overall sentiment, TSA (aka fine-grained sentiment analysis) is crucial for understanding how entities are portrayed in news articles. \citet{cortis_semeval-2017_2017} apply TSA on financial headlines, where sentiment is less implicit and topically constrained.  \citet{dufraisse_mad-tsc:_2023} and \citet{steinberger_multilingual_2011} present multilingual datasets for TSA in news articles. \citet{hamborg_newsmtsc:_2021} present a dataset for TSA on English news articles reporting on political topics, while \citep{balahur_sentiment_2013} focus on quotes from news articles. 
Overcoming the need for a labeled dataset, LLMs present as a possible solution for TSA due to their ICL abilities and broad background. \citet{huang_reducing_2020} conducted an analysis to identify and mitigate the entity bias of LLMs trained for sentiment analysis on Wikipedia and news articles. \citet{chumakov_generative_2023} leverage both few-shot learning and fine-tuning with GPT models on mixed-domain Russian and English datasets to model sentiment effectively without domain-specific data.

\section{Datasets and Models}

Our experiments utilize, to our knowledge, the only two available datasets for TSA in general news headlines, alongside one domain-specific dataset. These datasets cover four languages and employ different annotation styles.

\paragraph{STONE.}
%Filling the void in TSA for low-resource languages, 
The \stone dataset \cite{baric_target_2023} offers overall sentiment and targeted sentiment annotations for Croatian news headlines, using ternary labels (positive, neutral, negative). Each of the 2855 headlines has 6 labels assigned by 6 annotators, with an IAA of $\kappa$ = 0.416 (moderate agreement). Annotators were instructed using prescriptive, detailed guidelines (obtained from the authors upon our request).
%including many examples provided in full by the authors upon request.
%, with the highest level aligning with annotation guidelines and including provided examples.

\paragraph{SEN.}
The \sen \cite{baraniak_dataset_2021} dataset includes 3819 English and Polish news headlines. It comprises a Polish part (SEN\_pl), an English part (SEN\_en\_r) annotated by volunteer researchers, and an English part (SEN\_en\_amt) annotated using Amazon Mechanical Turk. The reported Fleiss' kappa IAA are $\kappa$ = .459, $\kappa$ = .309, and $\kappa$ = .303, respectively. Unlike \stone, \sen lacks raw labels, providing only an aggregated gold label per headline (positive, neutral, and negative), and was annotated using vaguer annotator guidelines, adhering more to the descriptive paradigm.

\paragraph{Spanish.} 
The Spanish dataset of \citet{salgueiro2022spanish} comprises 1976 headlines concerning the 2019 Argentinian Presidential Elections. Three annotators assigned ternary labels to each headline with masked targets, with IAA of $\alpha = .62$. The authors do not disclose annotation guidelines, which suggests the straight-forward descriptive paradigm.

\paragraph{Models.} 
We experiment with four open-source models: Neural Chat (NC) (7B), Llama 3 (8B), Phi-3 (3.8B), and Gemma (9B), pitted against two proprietary OpenAI models -- GPT-4 Turbo (560B) and GPT-3.5 Turbo (175B) \citep{openai2023gpt4} %\footnote{We ommit 'Turbo' in further text for brevity.}. 
(cf.~Appendix~\ref{sec:appendix} for more details).

\section{Experiments and Results}

\subsection{Predictive Accuracy}

We first evaluate the LLMs' accuracy of TSA on headlines and compare them to top-performing BERT* models. We use the BERT models specifically pre-trained for each language -- RoBERTa-base \cite{liu2019roberta} for English, BERTić \cite{ljubesic_bertic_2021} for Croatian, and Polish-RoBERTa-base-v2 \cite{dadas2020pretraining} for Polish -- and fine-tune each for TSA on the corresponding training set. For LLMs, we use zero-shot prompting on the test set, using basic prompts outlining the task and the target classes (cf.~Appendix~\ref{sec:appendix}).

\begin{table}
\centering
    {
    \small
    \begin{tabular}{lccccc} 
     \toprule
     & & & \multicolumn{3}{c}{\textbf{SEN}} \\
    \cmidrule(lr){4-6}
    & \textbf{STONE} & \textbf{SD} & en\_amt & en\_r & pl \\ 
     \midrule
     GPT 3.5 & 61.3 & 64.2 & 66.1 & 61.5 & 60.0\\ 
     GPT 4 & 65.9 & \textbf{67.0} & \textbf{68.8} & 63.2 & \textbf{69.5} \\
     % Mistral & 43.0 & 50.0 & 56.1 & 45.8 & 47.7\\
     Neural Chat & 59.8 & 63.0 & 66.3 & \textbf{63.8} & 58.1\\
     Llama 3 & 53.5 & 60.5 & 59.2 & 52.7 & 51.2\\
     Phi-3 & 43.5 & 61.7 & 58.3 & 52.7 & 47.3\\
     Gemma & 48.4 & 60.5 & 60.0 & 52.7 & 51.2\\
     BERT* & \textbf{74.9} & 66.7 & 63.6 & 56.2 & 61.9\\ 
    \hline
    \end{tabular}}
\caption{F1 scores across languages and datasets}
\label{tab:rezultati1}
\end{table}

Table~\ref{tab:rezultati1} presents the F1 scores on the test set portions of each dataset. On the descriptive datasets (SEN and Spanish), LLMs outperform BERT-based models. GPT-4 achieves the highest F1 score on the Polish \sen and the crowdsourced English \sen. Interestingly, on the English \sen annotated by researchers, NC outperforms both fine-tuned BERT models and GPT. However, on \stone -- the prescriptively annotated dataset -- BERTić surpasses all other models by a significant margin.
%
%Although the datasets differ in many aspects, 
We argue this performance difference might stem from using different annotation paradigms. The best-performing LLMs seem to grasp the descriptive paradigm well, performing TSA closest to annotators. On the other hand, the performance gap observed in LLMs on \stone may stem from the prompts' vagueness and lack of alignment with its prescriptiveness -- a question we explore next.

%This could be due to the highly prescriptive nature of the annotator guidelines, which dictate a specific interpretation of headline sentiment, a characteristic captured well by BERTić during fine-tuning. The performance gap observed in LLMs on \stone may stem from the prompts' vagueness and lack of alignment with that level of prescriptiveness -- a question we explore next.

\subsection{Level of Prompt Prescriptiveness}

We utilize the \stone dataset and its annotator guidelines to create six prompts of increasing prescriptiveness level, with each subsequent level incorporating additional information from the guidelines. Table~\ref{tab:levels} outlines these six levels (cf.~Appendix \ref{sec:appendix} for full prompts). Our goal is to assess the ability of LLMs to follow instructions as accurately as human annotators and to determine the most effective level of prompt prescriptiveness.

\begin{table}
    \centering
    {\small
    \begin{tabular}{c p{6 cm}} 
     \toprule
     \textbf{Level} & \textbf{Description} \\ 
     \midrule
    1 & Concise, exploring the fundamental concepts of sentiment and targeted sentiment.\\
     2 & Includes a definition of targeted sentiment specifically within the framework of news headlines. \\
    3 & Provided with concise guidelines. \\
     4 & Comprehensive instructions provided as guidelines, excluding examples. \\
    5 & Comprehensive instructions presented as guidelines, including examples and brief explanations. \\
     6 & Comprehensive instructions provided exactly as they were presented to the annotators.\\ 
     \bottomrule
    \end{tabular}}
    \captionof{table}{Short descriptions of prompt prescriptiveness levels (cf.~Appendix \ref{sec:appendix} for full prompts)\label{tab:levels}}
    %Level 1 contains basic instructions, while level 6 adheres to the prescriptive paradigm and is identical to annotator guidelines 
\end{table}

Table~\ref{tab:F1-levels} shows the results. We observe variance in performance across all levels for all models. GPT-4 consistently outperforms other models across all levels, with GPT-4 and Neural Chat reaching their performance peaks at level 4 (detailed instructions formatted as guidelines without examples) and GPT 3.5 performing best at level 3 (concise guidelines). The performance drop seen at levels 5 and 6, the only ones with few-shot examples, may be due to the sensitivity regarding the selection and ordering of examples, a phenomenon observed in few-shot prompting \citep{lu2022fantastically, chang-jia-2023-data}. The increasing accuracy from levels 1 to 4 suggests that more prescriptive instructions positively impact LLM performance. 
Despite their overall lower performance, Llama 3, Gemma, and Phi-3 significantly improve at levels 5 and 6 (few-shot prompts). This difference in performance could be due to instruction tuning, which may have reduced sensitivity to few-shot configuration and improved context following.
%\alert{While fine-tuned BERT still outperforms prompting at all prescriptiveness levels, the attractiveness of using LLM for TSA on headlines lies in their universality across languages and lack of need for labeled datasets.}

\begin{table}
\centering
\begin{adjustbox}{width=\columnwidth}
{\small
\begin{tabular}{ccccccc}
\toprule
  \textbf{Level} & NC & GPT 3.5 & GPT 4 & Llama 3 & Gemma & Phi-3  \\ 
 \midrule
 1 & 59.8 & 60.1 & 65.9 & 53.5 & 48.4 & 43.5 \\
 2 & 61.2 & 58.3 & 64.3 & 50.9 & 48.8 & 40.6\\
 3 & 61.5 & \textbf{65.7} & 69.9 & 52.9 & 55.8 & 44.2 \\
 4 & \textbf{63.1} & 64.0 & \textbf{70.2} & 51.9 &  53.1 & 43.6\\
 5 & 60.5 & 63.0 & 66.8 & 60.6 &  59.3& \textbf{49.4} \\
 6 & 62.5 & 64.5 & 68.2 &\textbf{ 61.9} &\textbf{ 59.4} & 46.3\\
 \hline
\end{tabular}}
\end{adjustbox}
\caption{F1 scores for levels of prompt prescriptiveness\label{tab:F1-levels}}
\end{table}

\subsection{Uncertainty Quantification}

Given the inherent subjectivity of TSA and leveraging the stochastic nature of predictions generated by LLMs, we explore how LLMs can model human label variation and whether this varies across levels of prompt prescriptiveness. Using \stone, we approach this question from two angles: (1) examining the relationship between LLMs' predictive and calibration accuracies and (2) investigating if the uncertainty of LLM predictions aligns with inter-annotator disagreement.

We use three uncertainty quantification methods: self-consistency sampling, distribution prompting, and verbal confidence assessment. 
\emph{Self-consistency sampling} (SCS) \citep{xiong_can_2023} leverages the inherent stochasticity of LLMs, influenced by internal parameters such as temperature. For each headline, we prompt the same model six times and accumulate the responses to mimic the distribution of six annotator responses, setting the temperature to 0.7 for all models. The second method, which we refer to as \emph{distribution prompting} (DP), prompts the model to explicitly predict how six annotators would label the targeted sentiment, directly resulting in a distribution of positive, neutral, and negative responses. Lastly, the \emph{verbal confidence assessment} (VCA) method \citep{xiong_can_2023} prompts the LLM to produce three predictions for each headline, representing each sentiment class, along with a confidence score ranging from 0 to 100.

%Figure \ref{fig:correlation} shows the correlations between label distribution entropies from LLMs and annotators across prompt prescriptiveness levels. While correlations vary, they are generally weak, indicating misalignment between LLM predictive uncertainty and human subjectivity. The highest correlation occurs with GPT 3.5 and DS at level 3.

\begin{figure}[t]
\centering
\hspace*{-0.5cm} 
\includegraphics[width=0.5\textwidth]{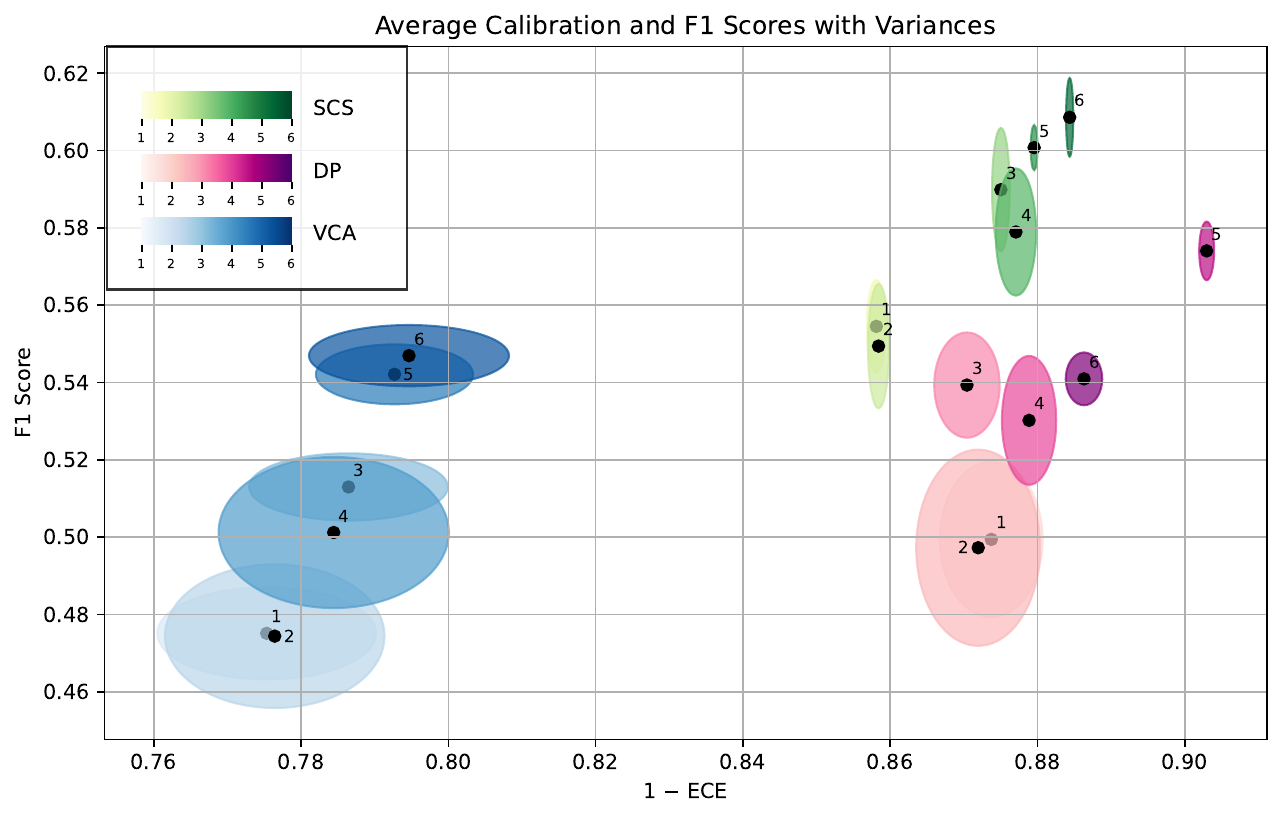}
\caption{Average F1 scores and calibration accuracy for uncertainty quantification methods across levels (shaded ellipses indicate covariances)
%The gray lines indicate the Pareto front.
}
\label{fig:f1-calibration}
\end{figure}

To analyze the calibration error, we consider only the labels with the highest confidence score for each headline and calculate the expected calibration error (ECE; cf.~Appendix \ref{ece}). Figure~\ref{fig:f1-calibration} compares the LLMs predictive accuracy (F1 score) against calibration accuracy, defined as 1-ECE, with both metrics averaged across all models (cf.~Tables~\ref{tab:F1-levels-complete} and \ref{tab:calibration-error} and Figure~\ref{fig:f1-calibration-complete} in Appendix~\ref{sec:results} for comprehensive data across all models).
As expected, as models' predictions are evaluated solely against the gold labels, F1 is higher for SCS than for DP and VCA. 
Average calibration accuracy is generally high (above $0.75$) across models and uncertainty methods. Higher prescriptiveness levels show an increasing trend in both predictive and calibration accuracy, with the optimal level varying by uncertainty method (Level 6 for SCS and VCA, and Level 5 for DP). This suggests that using prescriptive annotation guidelines can enhance LLM performance for prescriptive datasets.

%To analyze the calibration error, we consider only the labels with the highest confidence score for each headline and calculate the expected calibration error (ECE; cf.~Appendix \ref{ece}). Figure~\ref{fig:f1-calibration} compares the LLMs predictive accuracy (F1 score) against calibration accuracy, defined as 1-ECE, with prompt prescriptiveness level indicated by dot size \new{(cf. Tables ~\ref{tab:F1-levels-complete} and \ref{tab:Calibration-scores} for data)} As expected, as models' predictions are evaluated solely against the gold labels, F1 is higher for SCS than for DP and VCA. Calibration accuracy is generally high (above $0.6$) across models and uncertainty methods. Similar to predictive accuracy, calibration accuracy tends to increase with prescriptivity level, except for DP. The optimal configurations lie on the Pareto front, achieving a balance between F1 scores and calibration accuracy. 
%Among these, GPT-4 at prescriptiveness level 4 and Neural Chat at level 4 maximize the F1 score and calibration accuracy, respectively. GPT-3.5 at level 4 emerges as the best choice, achieving a balance between predictive and calibration accuracy.

\begin{figure}[t]
\centering
% \hspace*{-0.8cm} 
\includegraphics[width=0.4\textwidth]{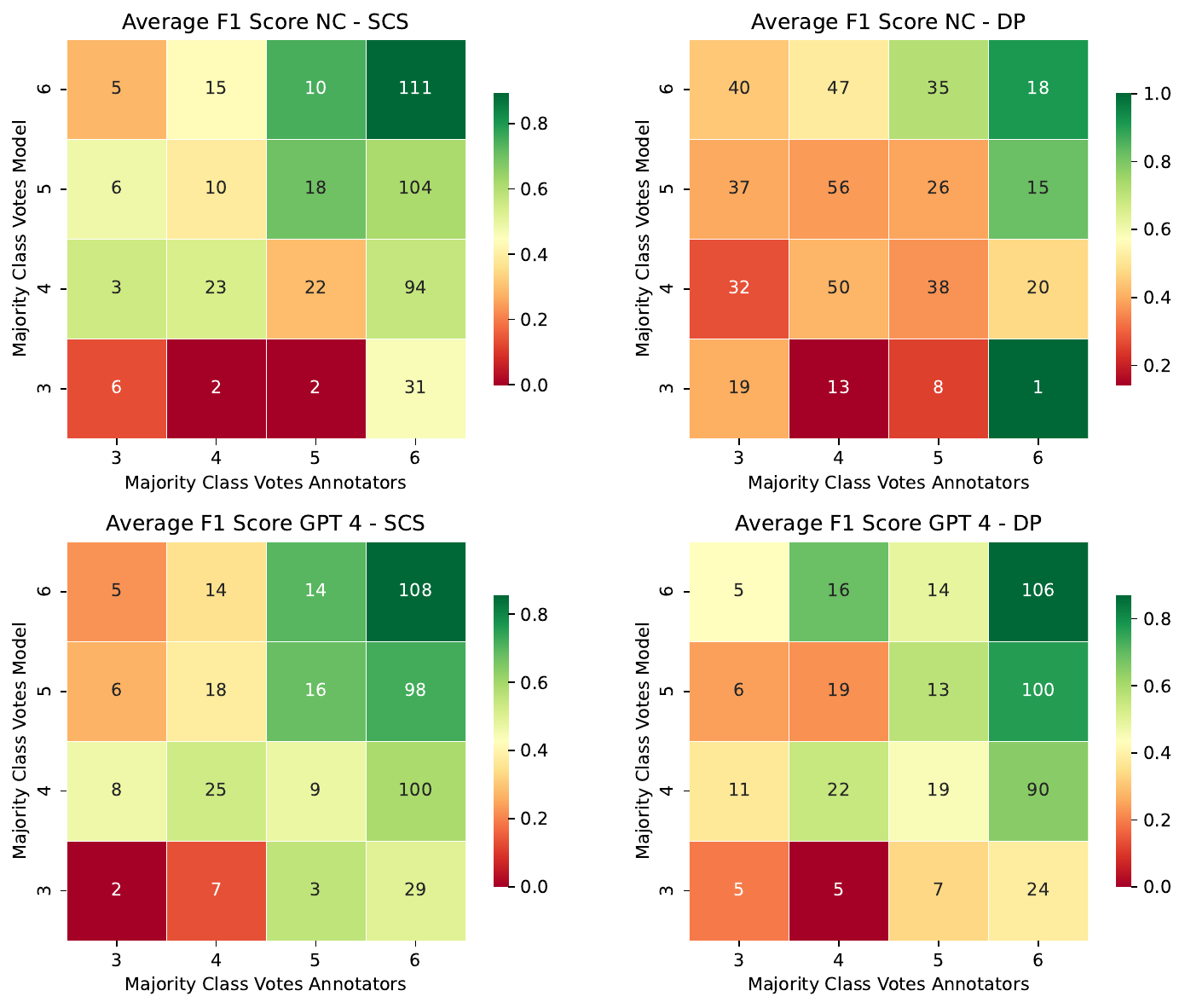}
%\caption{F1 score per majority vote bins for annotators (X) and model (Y) for SCS vs. DP for best open-source (NC) and best GPT variant.}
\caption{F1 score per majority vote bins for annotators (X) and model (Y) for SCS vs. DP for NC and GPT-4}
\label{fig:heatmap}
\end{figure}

Besides quantifying uncertainty, SCS and DP  can model human label variation, implicitly (SCS) or explicitly (DP). We compare these label distributions to human label variation. Figure~\ref{fig:heatmap} shows heatmaps of average F1 scores for the two best-performing open- and closed-source models, GPT-4 and NC. The axes represent the majority vote per instance by annotators and model. The highest F1 score is achieved when both the annotators' votes and the models' prediction are unanimous (6 votes). The lowest F1 scores are generally achieved for instances with less agreement within annotators or model votes. For GPT-4, DP performs similarly to SCS, whereas for NC, there is a significant performance drop and dispersion of model votes in bins, signaling the model is not grasping the concept. This suggests that SCS is a better choice for modeling label distribution across models.

\section{Conclusion}

Building on parallels with annotation paradigms for subjective tasks, we investigated the performance of LLM in-context learning for targeted sentiment analysis on news headlines. Our findings indicate that predictive accuracy increases with prompt prescriptiveness, though the optimal level varies by model, and only some models benefit from few-shot prompting. Calibration generally improves with prompt prescriptiveness, and self-consistency sampling aligns best with human label variation.

\section*{Limitations and Risks}

\paragraph{Limitations.}
We find several limitations in this work. Firstly, our choice of LLMs is restricted. This is primarily due to computing and budget constraints. We are aware that a more expansive collection of models is necessary for a more comprehensive overview of LLM performance, along with open-source models larger than 8B parameters. Additionally, we prompted both GPT models using batches of data, which impacted performance during initial tests, but did not warrant the high costs of repeating the prompt for each individual instance.

Secondly, the aspect of varying prescriptiveness in prompts was only evaluated on one dataset, \stone. To our knowledge, there are currently no publicly available datasets on TSA in news headlines annotated with detailed guidelines. Furthermore, since the dataset in focus is in Croatian, it is unclear whether a difference in performance is due to the difference in the ability for sentiment analysis or the general understanding of the language and its cultural and political background, both essential for the task.

Finally, while evaluating the effect of prompt prescriptiveness level, the six levels were chosen arbitrarily so that they resemble a logical step-up in detail level. This number and method of prompt generation can differ based on the task at hand and annotation guidelines.

\paragraph{Risks.}
The risks in our work are mostly connected with the risks associated with sentiment analysis. Automatically evaluating sentiment might promote exclusion towards certain entities. As we performed no masking of entities, internal model biases could affect the classification.

\bibliography{anthology,custom}

\appendix

\section{Appendix}
\label{sec:appendix}

\subsection{Additional Info on Datasets}

In Figure \ref{tab:datasets} the dataset sizes are shown. For the \stone dataset, we used the split for train, validation, and test sets given by the authors \citep{baric_target_2023}. For the variations of the \sen dataset, we used a 60/20/20 split generated using the sci-kit learn library with a fixed random seed of 42. 

\begin{table}
\centering
    {
    \small
    \begin{tabular}{lcccc} 
     \toprule
     & & \multicolumn{3}{c}{\textbf{SEN}} \\
    \cmidrule(lr){3-5}
    & \textbf{STONE} & en\_amt & en\_r & pl \\ 
     \midrule
     train set size & 1614  & 806 & 662 & 688 \\
     validation set size & 231 & 269 & 220 & 230 \\
     test set size & 462 & 269 & 220 & 230 \\
    \hline
    \end{tabular}}
\caption{Dataset sizes used in experiments}
\label{tab:datasets}
\end{table}

\begin{table}
\centering
    {
    \small
    \begin{tabular}{lcccc} 
     \toprule
     & & \multicolumn{3}{c}{\textbf{SEN}} \\
    \cmidrule(lr){3-5}
    & \textbf{STONE} & en\_amt & en\_r & pl \\ 
     \midrule
     learning rate & 1e-5  & 2e-5 & 2e-5 & 3e-5 \\
     batch size & 16 & 16 & 16 & 64 \\
     epoch size & 4 & 3 & 3 & 5 \\
    \hline
    \end{tabular}}
\caption{Optimal hyperparameters determined for each dataset: for STONE, the results are obtained using the BERTić model; for SEN\_en\_amt and SEN\_en\_r, RoBERTa-base is utilized; and for SEN\_pl, Polish-RoBERTa-base-v2 is employed.}
\label{tab:hyperparameters}
\end{table}

\subsection{Additional Information on Models}

For the BERT* models, we performed grid search for hyperparameter selection. We varied the learning rate, batch sizes and number of epochs. The optimal hyperparameters are shown in Table \ref{tab:hyperparameters}.

In our experiments, we employed the following LLMs:

    \paragraph{Neural Chat}  \textsuperscript{\footnote{https://ollama.com/library/neural-chat}} (7B): A fine-tuned model based on Mistral\textsuperscript{\footnote{https://mistral.ai/news/announcing-mistral-7b/}} with good coverage of domain and language.
    \paragraph{Llama 3 instruct (8B)}\textsuperscript{\footnote{https://ollama.com/library/llama3:instruct}}  : Instruction-tuned models fine-tuned and optimized for dialogue/chat use cases that outperform many of the available open-source chat models on common benchmarks.
    \paragraph{Phi-3 Mini instruct}\textsuperscript{\footnote{https://ollama.com/library/phi3:instruct/blobs/fa8235e5b48f}} (3.8B): Phi-3 Mini is a lightweight, state-of-the-art open model by Microsoft\textsuperscript{\footnote{https://www.microsoft.com/}}, trained with a focus on high-quality and reasoning dense properties.
    
    \paragraph{Gemma}\textsuperscript{\footnote{ https://ollama.com/library/gemma:instruct}} (8.5B): Gemma is a lightweight, state-of-the-art open model built by Google DeepMind.\textsuperscript{\footnote{https://deepmind.google/}}

    \paragraph{GPT-4 Turbo}\textsuperscript{\footnote{https://platform.openai.com/docs/models/gpt-4-turbo-and-gpt-4}} (560B): Latest generation OpenAI\textsuperscript{\footnote{https://openai.com/}} model in time of running our experiments. We used the \texttt{gpt-4-1106-preview} model. 
    
    \paragraph{GPT-3.5 }Turbo\textsuperscript{\footnote{https://platform.openai.com/docs/models/gpt-3-5-turbo}} (175B): Released in 2023, faster and more affordable OpenAI model. We used the \texttt{gpt-3.5-turbo-0125} model.

\subsection{Additional Information on GPU Usage}

We utilized a total of 201 hours of GPU resources. Specifically, 14 hours were allocated for obtaining results for optimal models and hyperparameters for BERT-based models. Additionally, 38 hours were dedicated to GPT 3.5 Turbo, 76 hours to GPT 4 Turbo, 62 hours to Neural Chat inference, and 11 hours to Mistral. Neural Chat and Mistral were run locally, while the GPT models were executed using the OpenAI Platform \footnote{https://platform.openai.com/docs/introduction}.

\subsection{Additional Information on Used Toolkits}

For tokenizing data to obtain results on BERT-based models, we utilized the PyTorch Transformers library\textsuperscript{\footnote{https://pytorch.org/hub/huggingface\_pytorch-transformers/}}.

\subsection{Complete results} \label{sec:results}
\begin{figure}[t]
\centering
% \hspace*{-0.8cm} 
\includegraphics[width=0.48\textwidth]{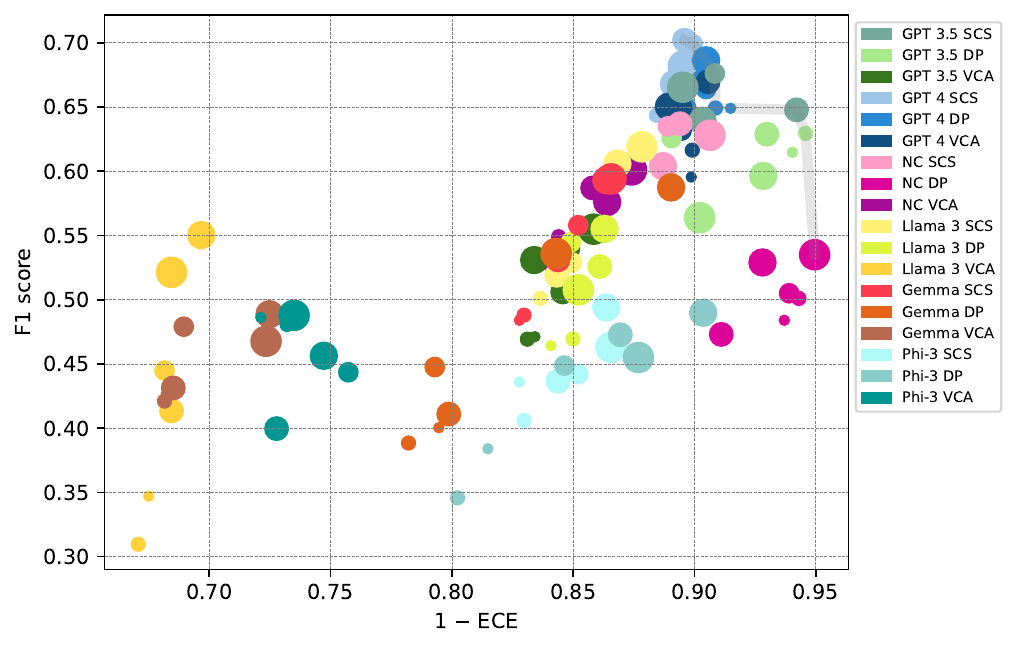}
\caption{Comparison of F1 scores and calibration accuracy for various uncertainty quantification methods and across levels of prompt prescriptiveness (indicated by dot size). The gray lines indicate the Pareto front.
}
\label{fig:f1-calibration-complete}
\end{figure}

In this section, we present complete results for the prescriptiveness prompting. In Table \ref{tab:F1-levels-complete} F1 scores are provided for all models and levels, and in Table \ref{tab:calibration-error} ECE is given per level and model. Figure \ref{fig:f1-calibration-complete} shows it graphically.

\begin{table*}
    \centering
% \begin{adjustbox}{width=\textwidth}
{\scriptsize
\begin{tabular}{ccccccccccccccccccc}
\toprule
\multirow{2}{*}{\textbf{Level}} & \multicolumn{3}{c}{\textbf{NC}} & \multicolumn{3}{c}{\textbf{GPT 3.5}} & \multicolumn{3}{c}{\textbf{GPT 4}} & \multicolumn{3}{c}{\textbf{Llama 3}} & \multicolumn{3}{c}{\textbf{Gemma}} & \multicolumn{3}{c}{\textbf{Phi-3}} \\
\cmidrule(lr){2-4} \cmidrule(lr){5-7} \cmidrule(lr){8-10} \cmidrule(lr){11-13} \cmidrule(lr){14-16} \cmidrule(lr){17-19}
 & SCS & DP & VCA & SCS & DP & VCA & SCS & DP & VCA & SCS & DP & VCA & SCS & DP & VCA & SCS & DP & VCA \\
\midrule
1 & 60.3 & 48.4 & 52.4 & 60.9 & 61.5 & 47.1 & 66.0 & 64.9 & 59.5 & 53.4 & 46.4 & 34.7 & 48.4 & 40.0 & 42.7 & 43.6 & 38.4 & 48.6 \\
2 & 62.0 & 50.1 & 54.9 & 61.4 & \textbf{62.9} & 46.9 & 64.4 & 64.9 & 61.6 & 50.1 & 46.9 & 30.9 & 48.8 & 38.8 & 42.1 & 40.6 & 34.6 & 48.1 \\
3 & 63.5 & 50.5 & 53.9 & \textbf{67.6} & 62.5 & 53.9 & 69.9 & 66.4 & 63.2 & 52.9 & 54.5 & 44.4 & 55.8 & 44.7 & 47.9 & 44.2 & 44.9 & 44.4 \\
4 & \textbf{63.7} & 47.3 & 58.7 & 64.8 & 62.8 & 50.6 & \textbf{70.2} & 66.9 & \textbf{66.9} & 51.9 & 52.6 & 41.3 & 53.1 & 41.1 & 43.1 & 43.6 & 47.2 & 39.9 \\
5 & 60.4 & 52.9 & 57.6 & 63.9 & 59.6 & 53.1 & 66.8 & \textbf{68.6} & 65.0 & 60.6 & \textbf{55.5} & \textbf{55.0} & 59.3 &\textbf{ 58.7} &\textbf{ 48.9} & \textbf{49.4 }&\textbf{ 48.9} & 45.6 \\
6 & 62.8 & \textbf{53.5} & \textbf{60.1} & 66.5 & 56.3 & \textbf{55.4} & 68.2 & 64.7 & 64.9 & \textbf{61.9} & 50.7 & 52.1 & \textbf{59.4} & 53.5 & 46.7 & 46.3 & 45.5 & \textbf{48.8 }\\
\bottomrule
\end{tabular}}
% \end{adjustbox}
\caption{F1 scores for levels of detail in prompt and uncertainty quantification metrics.}
\label{tab:F1-levels-complete}
\end{table*}

\begin{table*}
    \centering
% \begin{adjustbox}{width=\textwidth}
{\scriptsize
\begin{tabular}{ccccccccccccccccccc}
\toprule
\multirow{2}{*}{\textbf{Level}} & \multicolumn{3}{c}{\textbf{NC}} & \multicolumn{3}{c}{\textbf{GPT 3.5}} & \multicolumn{3}{c}{\textbf{GPT 4}} & \multicolumn{3}{c}{\textbf{Llama 3}} & \multicolumn{3}{c}{\textbf{Gemma}} & \multicolumn{3}{c}{\textbf{Phi-3}} \\
\cmidrule(lr){2-4} \cmidrule(lr){5-7} \cmidrule(lr){8-10} \cmidrule(lr){11-13} \cmidrule(lr){14-16} \cmidrule(lr){17-19}
 & SCS & DP & VCA & SCS & DP & VCA & SCS & DP & VCA & SCS & DP & VCA & SCS & DP & VCA & SCS & DP & VCA \\
\midrule
1 & 13.1 & 6.3 & 15.9 & 11.3 & 6.0 & 16.6 & 11.1 & \textbf{8.5} & 10.1 & 15.3 & 15.9 & 32.5 & 17.2 & 20.5 & 31.8 & 17.2 & 18.5 & 27.9 \\
2 & 12.1 & 5.7 & 15.6 & 10.9 & \textbf{5.4} & 16.9 & 11.6 & 9.1 & 10.1 & 16.4 & 15.0 & 32.9 & 17.0 & 21.8 & 31.9 & 17.0 & 19.8 & 26.8 \\
3 & 11.1 & 6.1 & 15.3 & 9.2 & 10.9 & 15.1 & \textbf{10.1} & 9.5 & 10.5 & 15.1 & 15.1 & 31.8 & 14.8 & 20.7 & 31.1 & 14.8 & 15.4 & \textbf{24.3} \\
4 & 10.6 & 8.9 & 14.2 & \textbf{5.8} & 7.0 & 15.4 & 10.4 & 9.7 & \textbf{9.4} & 15.7 & 13.9 & 31.6 & 15.6 & 20.1 & 31.5 & 15.6 & 13.1 & 27.2 \\
5 & 11.3 & 7.2 & 13.6 & 9.7 & 7.2 & 16.6 & 10.8 & 9.5 & 11.1 & 13.2 & \textbf{13.7} & \textbf{30.3} & 13.6 & \textbf{11.0} & \textbf{27.5} & 13.6 & \textbf{9.6} & 25.3 \\
6 & \textbf{9.4} & \textbf{5.1} & \textbf{12.6} & 10.5 & 9.8 & \textbf{14.2} & 10.5 & 10.6 & 10.7 & \textbf{12.2} & 14.8 & 31.6 & \textbf{13.4} & 15.7 & 27.7 & \textbf{13.4} & 12.3 & 26.5 \\
\bottomrule
\end{tabular}}
% \end{adjustbox}
\caption{Expected calibration error (ECE) for levels of detail in prompt and uncertainty quantification metrics.}
\label{tab:calibration-error}
\end{table*}

\subsection{Measurements}

\textbf{\textit{Expected calibration error (ECE)}}
\label{ece}
Model predictions are categorized into $m$ quantile-scaled bins $B_i$, choosing $m = 10$ for this analysis. For each bin, we calculate both the average accuracy $\text{acc}(B_i)$ and average uncertainty $\text{uncert}(B_i)$. The Expected Calibration Error (ECE) is then derived as the weighted sum of the absolute differences between these averages, with weights proportional to the bin size $n$. A lower ECE indicates superior model calibration, formalized by the equation:

\begin{equation}
    \text{ECE} = \sum_{i=1}^{m} \frac{|B_i|}{n} \left| \text{acc}(B_i) - \text{uncert}(B_i) \right|.
\end{equation}

\subsection{Prescriptiveness Levels Used in Prompts}
In this section, the complete prompts system and user prompts are given in tables. The yellow highlight shows an expansion in text and information compared to the previous level.

\begin{table*}
    \centering
    {\small
    \begin{tabular}{c c} 
     \toprule
     \textbf{Level} & \textbf{Prompt} \\ 
     \midrule
    1 & \parbox{\textwidth}{You are a helpful assistant who performs targeted sentiment classification in Croatian news headlines. The available sentiment classes are positive, neutral, and negative. For each given headline, identify the targeted sentiment class towards the entity.}\label{dodaci:upute_lvl_1} \\
    \midrule
    2 & \parbox{\textwidth}{You are a helpful assistant who performs targeted sentiment classification in Croatian news headlines. 
\hl{Targeted sentiment involves understanding the author's intention to evoke emotion towards a target entity, considering the deliberate choice in conveying news and recognizing that the same information can be presented in various ways, with the understanding that such intentional choices aid in detecting the targeted sentiment.} The available sentiment classes are positive, neutral, and negative. For each given headline, identify the targeted sentiment class towards the entity.}\label{dodaci:upute_lvl_2}\\
    \midrule
    3 & \parbox{\textwidth}{
    You are a helpful assistant who performs targeted sentiment classification in Croatian news headlines.
\hl{Targeted sentiment is the emotional stance the author aims to convey specifically towards a mentioned entity. It involves interpreting the intention behind the author's choice of language and tone when discussing the target entity. The sentiment is not only influenced by the conveyed information but also by the author's subjective evaluation and emotional coloring of the entity. Actions associated with the entity play a role in determining the sentiment, with negative actions implying a negative quality and, consequently, a negative sentiment. Distinguishing between negative actions and negative occurrences is crucial, as negative occurrences towards the entity don't color the entity. In headlines featuring a quote, the entity authoring the quote is attributed neutral sentiment as they are merely conveying an opinion. The overall goal of the author, whether it be praise or criticism, is considered in cases of headlines with a mix of positive and negative views. In summary, targeted sentiment is the nuanced emotional evaluation directed specifically at a particular entity within the context of news reporting.}
The available sentiment classes are positive, neutral, and negative. For each given headline, identify the targeted sentiment class towards the entity.
} \\
\midrule
    \end{tabular}}
    \captionof{table}{System prompts used for inference on the \stone dataset.}
\label{tab:levels1}
\end{table*}

\begin{table*}
    \centering
    {\small
    \begin{tabular}{c c} 
     \toprule
     \textbf{Level} & \textbf{Prompt} \\ 
 \midrule
    4 & \parbox{\textwidth}{
    
You are a helpful assistant who performs targeted sentiment classification in Croatian news headlines. 
\\ \\
\hl{
Guidelines for Targeted Sentiment Annotation:
\\ \\
1.	Detecting Sentiment through Author's Intent and News Presentation:
Evaluate the intended sentiment towards an entity by analyzing the emotions the author aims to evoke and recognizing that news can be conveyed in multiple ways, with the chosen manner of conveyance serving a purpose and aiding in targeted sentiment detection.
\\ \\
2.	Impact of Entity Actions:
Acknowledge that entity actions influence sentiment, with negative actions implying negative quality. However, distinguish between negative actions undertaken by the entity and negative occurrences directed towards the entity that do not inherently portray the entity in a negative light.
\\ \\
3.	Neutrality of Quoting Authors:
In headlines featuring quotes, two types of entities are involved: the statement's author and the entities mentioned in the quote. If the target entities in the headline are the authors of the statement, the sentiment towards them typically leans towards neutrality because, in this scenario, they serve as conveyors of an opinion rather than direct subjects of sentiment.
\\ \\
4.	Overall Authorial Goal:
Consider the author's overall goal, whether it involves praise or criticism, especially in mixed-view headlines.
\\ \\}
The available sentiment classes are positive, neutral, and negative. For each given headline, identify the targeted sentiment class towards the entity.
}
\\
\midrule
    5 & \parbox{\textwidth}{
    
You are a helpful assistant who performs targeted sentiment classification in Croatian news headlines. 
\\\\
Guidelines for Targeted Sentiment Annotation:
\\\\
1.	Detecting Sentiment through Author's Intent and News Presentation:\\
Evaluate the intended sentiment towards an entity by analyzing the emotions the author aims to evoke and recognizing that news can be conveyed in multiple ways, with the chosen manner of conveyance serving a purpose and aiding in targeted sentiment detection.
\\
\hl{
Examples Illustrating Sentiment towards Entity Solin:\\\\
Headline: 'SRAMOTA USolinuse djeca nemaju gdje liječiti, roditelji očajni'\\
Targeted Sentiment: Negative\\
Explanation: The author criticizes Solin, suggesting a disgraceful situation where children lack medical care, portraying a negative sentiment.\\\\
Headline: 'U Solinu radi samo jedna pedijatrica, roditelji traže hitno rješenje'\\
Targeted Sentiment: Negative\\
Explanation: The negative sentiment persists as the author emphasizes the shortage of pediatricians in Solin, prompting urgent solutions according to parents.\\\\
Headline: 'U Solinu nastupio nedostatak liječničkog kadra'\\
Targeted Sentiment: Neutral\\
Explanation: The sentiment is neutral here as the author focuses on conveying information about the shortage of medical staff without explicitly criticizing the responsible institutions.
}}
    \end{tabular}}
    \captionof{table}{System prompts used for inference on the \stone dataset.}
\label{tab:levels2}
\end{table*}

\begin{table*}
    \centering
    {\small
    \begin{tabular}{c c} 
     \toprule
     \textbf{Level} & \textbf{Prompt} \\ 
 \midrule
    5 & \parbox{\textwidth}{
  
2.	Impact of Entity Actions:\\
Recognize that entity actions play a role in shaping sentiment, particularly with negative actions like murder and theft suggesting a negative quality. However, distinguish between negative actions where the entity is the perpetrator and negative occurrences where the entity is the recipient. Acknowledge that in the case of negative occurrences, the entity cannot be held responsible for the consequences but may be in a negative situation as a result, implying neutrality in the assessment.
\\
\hl{
Headlines with negative quality of entities linked to their actions:
\\

a) Examples of linking entity quality to actions:\\
Headline: 'Bivša tehnološka direktorica Elizabeth Holmes osuđena na 11 godina zatvora'\\
Entity: Elizabeth Holmes\\
Targeted Sentiment: Negative\\
Explanation: Negative sentiment is assigned to Elizabeth Holmes based on her negative actions.
\\\\
Headline: 'Zbog ubojstva srpskih civila sudit će se Đuri Brodarcu, bivšem Sanaderovom savjetniku'\\
Entity: Đuro Brodarac\\
Targeted Sentiment: Negative\\
Explanation: Negative sentiment is assigned to Đuro Brodarac due to his association with a serious crime.
\\

b) Examples of negative occurences towards the entity.
\\\\
Headline: 'Potres u Indoneziji: Poginulo najmanje 46 ljudi, ozlijeđenih oko 700'\\
Entity: Indonezija\\
Targeted Sentiment: Neutral\\
Explanation: Neutral sentiment is assigned to Indonesia as the entity is a recipient of a negative occurrence.
\\\\
Headline: 'Horor u Mogadišuu: U terorističkom napadu na hotel 10 mrtvih, ozlijeđen i somalijski ministar'\\
Entity: Mogadišu\\
Targeted Sentiment: Neutral\\
Explanation: Similar to the previous example, neutral sentiment is assigned to Mogadishu as it is a recipient of a negative occurrence.
\\

}
3.	Neutrality of Quoting Authors:\\
Define sentiment towards the entity by considering the author's stance in a statement, whether the author is the headline writer or the individual quoted. When conveying someone's sentiment in a quote, transfer that sentiment to the mentioned entity. In headlines quoting individuals, recognize two entity types: the statement's author and the entities mentioned in the quote. If the target entities in the headline are the authors of the statement, the sentiment is typically neutral since they serve as conveyors of an opinion.
\\
\hl{
Examples of Handling Quotes in Headlines:
\\\\
Headline: 'Milanović: Žao mi je što sam podržao Bidena'\\
Entity: Milanović\\
Targeted Sentiment: Neutral\\
Entity: Biden\\
Targeted Sentiment: Negative\\
Explanation: Neutral sentiment is assigned to Milanović, who is conveying an opinion, while negative sentiment is assigned to Biden based on the conveyed sentiment.
\\\\
Headline: 'Gotovac: Ako sam ja politički antitalent, onda je tom antitalentu išlo bolje nego Grbinu'\\
Entity: Gotovac\\
Targeted Sentiment: Positive\\
Entity: Grbin\\
Targeted Sentiment: Negative\\
Explanation: Positive sentiment is assigned to Gotovac, who comments on himself, while negative sentiment is assigned to Grbin based on the conveyed sentiment.
\\\\
 Headline: 'Anka Mrak Taritaš: Tužna sam i razočarana situacijom u Zagrebu. Tomašević ne bi dobio dobru ocjenu'\\
Entity: Anka Mrak Taritaš\\
Targeted Sentiment: Neutral\\
Entity: Tomašević\\
Targeted Sentiment: Negative\\
Explanation: Neutral sentiment is assigned to Anka Mrak Taritaš, the quoted individual, while negative sentiment is assigned to Tomašević based on the conveyed sentiment.
}
}
\\
\midrule
    \end{tabular}}
    \captionof{table}{System prompts used for inference on the \stone dataset.}
\label{tab:levels3}
\end{table*}

\begin{table*}
    \centering
    {\small
    \begin{tabular}{c c} 
     \toprule
     \textbf{Level} & \textbf{Prompt} \\ 
     \midrule
     5 & \parbox{\textwidth}{
     
4.	Overall Authorial Goal:\\
Consider the author's overall goal, whether it involves praise or criticism, especially in mixed-view headlines.
\\
\hl{
Example of a Combined Statement (Combination of Positive and Negative Views)
\\\\
Headline: 'Vanna je definitivno promijenila stil naglavačke i dosadne kombinacije zamijenila onima koje prate trendove'\\
Entity: Vanna\\
Targeted Sentiment: Positive\\
Explanation: A positive sentiment is attributed to Vanna because the author's intention is to praise the improvement in her style, despite simultaneously criticizing her previous dressing style.}
\\\\
The available sentiment classes are positive, neutral, and negative. For each given headline, identify the targeted sentiment class towards the entity.\\
}\label{dodaci:upute_lvl_1} \\
    \midrule
    6 & \parbox{\textwidth}{You are a helpful assistant who performs targeted sentiment classification in Croatian news headlines.  Here are some guidelines for detecting targeted sentiment in news headlines:

\hl{ To determine sentiment towards an entity, we consider the kind of emotion the statement's author intended to evoke regarding the target entity,
that is, how the author intended to "color" that entity. To aid in discerning the intended sentiment towards the entity, one can consider the fact that the
same piece of news can always be conveyed in multiple ways. The chosen manner of conveying a piece of news is selected with a purpose, and understanding
that intention can be utilized for targeted sentiment detection.}

  An example of various ways of reporting the same news about entity Solin:
\\\\
Headline: 'SRAMOTA USolinuse djeca nemaju gdje liječiti, roditelji očajni'\\
Targeted Sentiment: Negative\\
Explanation: Negative sentiment is attributed to Solin due to the author's intention to criticize the institution for the shortage of pediatricians.
\\\\
Headline: 'U Solinu radi samo jedna pedijatrica, roditelji traže hitno rješenje'\\
Targeted Sentiment: Negative\\
Explanation: Similar negative sentiment is conveyed towards Solin by criticizing the shortage of medical staff.
\\\\
Headline: 'U Solinu nastupio nedostatak liječničkog kadra'\\
Targeted Sentiment: Neutral\\
Explanation: Neutral sentiment is assigned as the author's intention is to convey information without criticizing the responsible institutions.
\\\\
\hl{When detecting targeted sentiment, we can assign a quality to the target entity as an aid in determining sentiment, based on the emotion the statement's
author associates with it. The quality of the entity is linked to the actions of that entity, which can be either negative or positive. Negative actions of the entity,
such as murder, theft, and other illegal or socially unacceptable activities like insults, are attributed to the quality of that entity. Negative actions signify
a negative quality of the entity, implying a negative sentiment. The same approach will be applied in cases of positive actions of the entity, indicating a positive sentiment towards the entity.
It is necessary to distinguish between the negative actions of an entity and negative occurrences towards the entity. In the case of negative actions by
the entity, the entity is the perpetrator and therefore responsible for that action. In the case of negative occurrences towards the entity,
the entity is the recipient of the negative action and cannot be held responsible for the consequences of the action, although it may be in a negative
situation as a result.}
\\\\
Examples of linking entity quality to actions:
\\\\
Headline: 'Bivša tehnološka direktorica Elizabeth Holmes osuđena na 11 godina zatvora'\\
Entity: Elizabeth Holmes\\
Targeted Sentiment: Negative\\
Explanation: Negative sentiment is assigned to Elizabeth Holmes based on her negative actions.
\\\\
\\}\label{dodaci:upute_lvl_2}\\
\midrule
    \end{tabular}}
    \captionof{table}{System prompts used for inference on the \stone dataset.}
\label{tab:ENV}
\end{table*}

\begin{table*}
    \centering
    {\small
    \begin{tabular}{c c} 
     \toprule
     \textbf{Level} & \textbf{Prompt} \\ 
    \midrule
    6 & \parbox{\textwidth}{Headline: 'Zbog ubojstva srpskih civila sudit će se Đuri Brodarcu, bivšem Sanaderovom savjetniku'\\
Entity: Đuro Brodarac\\
Targeted Sentiment: Negative\\
Explanation: Negative sentiment is assigned to Đuro Brodarac due to his association with a serious crime.
\\\\
Examples of negative occurences towards the entity.
\\\\
Headline: 'Potres u Indoneziji: Poginulo najmanje 46 ljudi, ozlijeđenih oko 700'\\
Entity: Indonezija\\
Targeted Sentiment: Neutral\\
Explanation: Neutral sentiment is assigned to Indonesia as the entity is a recipient of a negative occurrence.
\\\\
Headline: 'Horor u Mogadišuu: U terorističkom napadu na hotel 10 mrtvih, ozlijeđen i somalijski ministar'\\
Entity: Mogadišu\\
Targeted Sentiment: Neutral\\
Explanation: Similar to the previous example, neutral sentiment is assigned to Mogadishu as it is a recipient of a negative occurrence.
\\

\hl {  We define sentiment towards the entity as the author's stance towards the target entity in a statement. The statement's author can be the person
who wrote the article headline or the author whose quote is conveyed in the form of the article headline. When conveying someone's negative/positive sentiment in
a quote or paraphrase, that sentiment is transferred to the entity. In headlines conveying someone's quote, there are two types of entities - the statement's
author and the entities mentioned in the quote. If the target entities in the headline are the authors of the statement, the sentiment towards them will usually
be neutral because, in this case, they are just conveyors of an opinion. An exception is the following example with entity Gotovac, where the statement's author comments on himself, and the expressed sentiment is then transferred to the author himself.}
\\\\
Examples of Handling Quotes in Headlines:
\\\\
Headline: 'Milanović: Žao mi je što sam podržao Bidena'\\
Entity: Milanović\\
Targeted Sentiment: Neutral\\
Entity: Biden\\
Targeted Sentiment: Negative\\
Explanation: Neutral sentiment is assigned to Milanović, who is conveying an opinion, while negative sentiment is assigned to Biden based on the conveyed sentiment.
\\\\
Headline: 'Gotovac: Ako sam ja politički antitalent, onda je tom antitalentu išlo bolje nego Grbinu'\\
Entity: Gotovac\\
Targeted Sentiment: Positive\\
Entity: Grbin\\
Targeted Sentiment: Negative\\
Explanation: Positive sentiment is assigned to Gotovac, who comments on himself, while negative sentiment is assigned to Grbin based on the conveyed sentiment.
\\\\
 Headline: 'Anka Mrak Taritaš: Tužna sam i razočarana situacijom u Zagrebu. Tomašević ne bi dobio dobru ocjenu'\\
Entity: Anka Mrak Taritaš\\
Targeted Sentiment: Neutral\\
Entity: Tomašević\\
Targeted Sentiment: Negative\\
Explanation: Neutral sentiment is assigned to Anka Mrak Taritaš, the quoted individual, while negative sentiment is assigned to Tomašević based on the conveyed sentiment.
\\
\\}\label{dodaci:upute_lvl_2}\\
\midrule
    \end{tabular}}
    \captionof{table}{System prompts used for inference on the \stone dataset.}
\end{table*}

\begin{table*}
    \centering
    {\small
    \begin{tabular}{c c} 
     \toprule
     \textbf{Level} & \textbf{Prompt} \\ 
    \midrule
    6 & \parbox{\textwidth}{\hl{In the case of a headline containing a combination of positive and negative views towards the entity, the final goal of the author towards the
entity is considered, i.e., whether the author aimed for praise or criticism.}
\\

 Example of a Combined Statement (Combination of Positive and Negative Views):
\\

Headline: 'Vanna je definitivno promijenila stil naglavačke i dosadne kombinacije zamijenila onima koje prate trendove'\\
Entity: Vanna\\
Targeted Sentiment: Positive\\
Explanation: Positive sentiment is assigned to Vanna as the author's intention is to praise the improvement in her style despite also criticizing her previous dressing choices.
\\

The available sentiment classes are positive, neutral, and negative. For each given headline, identify the targeted sentiment class towards the entity.
\\
\\}\label{dodaci:upute_lvl_2}\\
\midrule
    \end{tabular}}
    \captionof{table}{System prompts used for inference on the \stone dataset.}
\end{table*}

\begin{table*}
    \centering
    {\small
    \begin{tabular}{c c} 
     \toprule
     \textbf{} & \textbf{Prompt} \\ 
    \midrule
    \textbf{SCS} & \parbox{\textwidth}{
    Classify targeted sentiment towards entity \{\textit{entity}\} in the following news headline: \{{\textit{headline}}\}
    }\label{dodaci:upute_lvl_2}\\
\midrule
 \textbf{DP} & \parbox{\textwidth}{
    Your task is to imagine you are representing 6 different people detecting the targeted sentiment in Croatian news headlines, each following the given guidelines.
  For a headline and an entity, you need to return detected targeted sentiment for each of the 6 voters.
  \\
  Detect targeted sentiment for entity '{\textit{entity}}' in headline: '{\textit{headline}}'. Possible sentiment classes are positive, neutral and negative.  Please return the answer in JSON format like:\\
   {["targeted sentiment 1":"class 1"\\
   "targeted sentiment 2":"class 2"\\
   "targeted sentiment 3":"class 3"\\
    "targeted sentiment 4":"class 4"\\
   "targeted sentiment 5":"class 5"\\
    "targeted sentiment 6":"class 6"]}
    }\label{dodaci:upute_lvl_2}\\
\midrule
 \textbf{VCA} & \parbox{\textwidth}{
    You are a helpful assistant who performs targeted sentiment classification in Croatian news headlines. Following the given guidelines, please return the confidence for detection of each class.
 \\ 
Detect targeted sentiment for entity \{\textit{entity}\} in headline: \{\textit{headline}\}. Possible sentiment classes are positive, neutral and negative. 
\\
Please return the confidence for each class in format like:\\
{["confidence positive class", "confidence neutral class" ,"confidence negative class"]}}
\\
\midrule
    \end{tabular}}
    \captionof{table}{User prompt used for inference on the \stone dataset accross methods for uncertainty quantification.}
\label{tab:ENV}
\end{table*}

\end{document}